\documentclass[a4paper,twoside]{article}
\pdfoutput=1
\usepackage{epsfig}
\usepackage{graphicx}
\usepackage{subcaption}
\usepackage{calc}
\usepackage{amssymb}
\usepackage{amstext}
\usepackage{amsmath}
\usepackage{amsthm}
\usepackage{multicol}
\usepackage{pslatex}
\usepackage{apalike}
\usepackage{comment}
\usepackage{hyperref}
\usepackage[bottom]{footmisc}
\usepackage{SCITEPRESS}     
\usepackage{xcolor} 
\definecolor{pink}{RGB}{218,0,205}
\definecolor{blue}{RGB}{33,255,255}
\definecolor{purple}{RGB}{19,0,255}

\begin{document}

\title{ALiSNet: \textbf{A}ccurate and \textbf{Li}ghtweight Human \textbf{S}egmentation \textbf{Net}work \\ for Fashion E-Commerce}


\author{\authorname{Amrollah Seifoddini\orcidAuthor{0000-0002-6757-3175}, Koen Vernooij, Timon Künzle, Alessandro Canopoli, Malte Alf, \newline Anna Volokitin and  Reza Shirvany}
\affiliation{Zalando SE}
\email{amrollah.seifoddini@zalando.ch, kvernooij4@gmail.com, \{timon.kuenzle, alessandro.canopoli, malte.alf, anna.volokitin\}@zalando.ch, reza.shirvany@zalando.de}
}


\abstract{Accurately estimating human body shape from photos can enable innovative applications in fashion, from mass customization, to size and fit recommendations and virtual try-on.
Body silhouettes calculated from user pictures are effective representations of the body shape for downstream tasks. Smartphones provide a convenient way for users to capture images of their body, and on-device image processing allows predicting body segmentation while protecting users' privacy. 
Existing off-the-shelf methods for human segmentation are closed source and cannot be specialized for our application of body shape and measurement estimation. Therefore, we create a new segmentation model by simplifying Semantic FPN with PointRend, an existing accurate model. We finetune this model on a high-quality dataset of humans in a restricted set of poses relevant for our application.
We obtain our final model, ALiSNet, with a size of 4MB and 97.6~$\pm$~1.0\% mIoU, compared to Apple Person Segmentation, which has an accuracy of 94.4~$\pm$~5.7\% mIoU on our dataset. 
}

\onecolumn \maketitle \normalsize \setcounter{footnote}{0} \vfill

\section{\uppercase{Introduction}}
\label{sec:introduction}


Human segmentation has emerged as foundational to applications across a diverse range from autonomous driving to social media, virtual and augmented reality, and online fashion.  
In the case of online fashion, giving users a way to easily capture their body shape is valuable, since it can be used to recommend appropriate clothing sizes or to enable virtual try-on. However, to determine the right size and fit of clothing, the body shape needs to be determined with very high accuracy in order to be of value. For example, in an image of $2$k resolution in height, a segmentation error of two pixels on the boundary can change a measurement such as chest circumference by $10$mm.
Users’ body shape can be more accurately determined if they wear tight-fitting clothes, making it even more important than in other applications to preserve privacy. Hence, mobile human segmentation is a good fit for fashion applications, as images can be both captured and processed on-device.  


%

\begin{figure}[!htbp]
\begin{center}
  \includegraphics[width=1.0\linewidth]{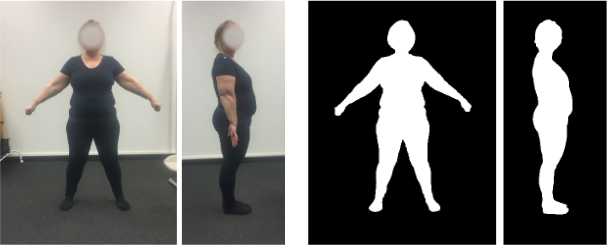}
\end{center}
  \caption{Ground truth body annotations. The boundary in particular is critical for body shape prediction.}
  \label{fig:optimass-example-gt}
\end{figure} 


In this paper we propose an approach to achieve an accurate and lightweight human segmentation method for these applications.
Although off-the-shelf mobile human segmentation methods are available, such as Apple Person Segmentation~\cite{apple_segmentation} and Google MLKit's BlazePose~\cite{blazepose}, these methods are closed-source and cannot be adapted to our task to achieve the required accuracy. Instead, we design a model based on Semantic FPN with PointRend for our task.

Crucial to the success of our method is finetuning on a task specific dataset of user-taken photos in front and side views, as shown in~\autoref{fig:optimass-example-gt}. Orthogonal views such as this are commonly used in various anthropometry setups, e.g.~\cite{smith2019towards}. Such silhouettes can be used to model the 3D body shape of users, as proposed in~\cite{dibra2016hs,dibra2017human,smith2019towards} or to directly predict measurements~\cite{yan2021silhouette} for fashion applications.
While relying on the large body of publicly available data for the segmentation task, we augment it by using a small yet specific dataset of $6147$ high resolution images with highly accurate annotations to overcome the limitations of publicly available data.


Our main contributions are thus two-fold: First, we demonstrate that a relatively small set of high-quality annotations can boost segmentation accuracy. 
Second, we simplify a large and high quality baseline method, Semantic FPN~\cite{Kirillov_2019_CVPR} with PointRend refinement~\cite{kirillov2020pointrend} with a few steps to achieve almost the same performance with 100$\times$ model size.  The main changes to the original model are: exchanging the backbone with a modified version of the mobile-optimized MnasNet~\cite{tan2019mnasnet}, using quantization-aware training, and removing network components that we found not to be contributing to segmentation accuracy.
Our final \textbf{A}ccurate and \textbf{Li}ghtweight mobile human \textbf{S}egmentation \textbf{Net}work (ALiSNet), achieves $97.6$\%~mIoU and is $4$MB in size, where an off-the-shelf method such as BlazePose-Segmentation achieves $93.7$\%~mIoU on our data, with a $6$MB model, and Apple Person Segmentation achieves $94.4$\% mIoU. It was not possible to fine-tune either of these models to our data as they are closed-sourced.
Additionally, ALiSNet's accuracy is only marginally lower than the $97.8$\%~mIoU achieved by the $350$~MB baseline.

\section{\uppercase{Related Work}}


The categories of methods most relevant to our work in the domain of on-device human segmentation are portrait editing, video call background effects, and general-purpose real time whole body segmentation methods.

Many portrait editing predict alpha mattes, which are masks that allow blending foreground and background regions. In this application, having accurate segmentation of textures such as wisps of hair is very important. Google Pixel's alpha matting method~\cite{google_alpha_matting} relies on data collected using a custom volumetric lighting setup. Apple Person Segmentation~\cite{apple_segmentation} in \emph{accurate} mode also belongs to this category of methods. However, such accuracy on the texture level is not necessary for our application. Besides, most existing alpha matting methods are trained only on faces.

Real-time portrait segmentation methods for video calls focus on segmenting the human upper body. ExtremeC3Net~\cite{park2019extremec3net} and SiNet~\cite{li2020fast} are examples of models that achieve very good performance under a parameter count of 200K. 

There are also several methods focused on real-time segmentation of the whole body. One example is Google MLKit BlazePose-Segmentation~\cite{blazepose} which relies on correct prediction of body bounding box.
\cite{strohmayer2021efficient} focuses on reducing latency for general purpose human segmentation. ~\cite{Liang2022} introduces Multi-domain TriSeNet Networks for the real-time single person segmentation for photo-editing applications. \cite{Xi2019} uses saliency map derived from accurate pose information to improve segmentation accuracy especially in multi-person scenes.

\cite{han2020ghostnet} categorizes the set of techniques for reducing model size into \emph{model compression} and \emph{compact model design} methods. Although we make use of quantization-aware training~\cite{wu2015quantized} in this paper, which is a compression technique, we mostly take advantage of compact model components. These include compact networks that can be used as feature extractors, such as MnasNet~\cite{tan2019mnasnet}, FBNet~\cite{wu2019fbnet} and  MobileNetv3~\cite{howard2019searching} which have been found using Neural Architecture Search. 

We recommend the related work section of ~\cite{knapp2021real} for a more extensive review of works related to mobile person segmentation.

Finally, our work can be used in downstream applications for estimating body shape. This is an active research area, with several approaches of estimating body shape from silhouette, such as~\cite{8360117,song20163d,ji2019human,dibra2016hs}.

\begin{figure}[!htbp]
\begin{center}
   \includegraphics[width=0.6\linewidth]{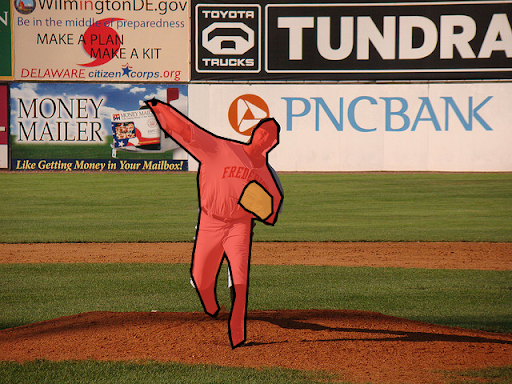}
\end{center}
   \caption{An example image and annotation from the COCO dataset. Note the low accuracy of the polygon annotation.}
\label{fig:coco-example}
\end{figure}

\paragraph{Datasets}
We review several datasets with permissive licenses that include person segmentation labels. These include MS COCO~\cite{coco}, shown in~\autoref{fig:coco-example}, LVIS~\cite{lvis} (contains higher quality annotations for COCO images), Google Open Images~\cite{google-open-images}. There are limitations with each of these datasets. COCO annotations are based on polygons and therefore not accurate around the object boundaries. LVIS annotations are very accurate and dense in each image but they are not yet available for the entire COCO dataset, especially in the human category where only around $1.8$k images are annotated. Finally, the Google Open Images dataset is sparse in segmentation coverage in each image compared to COCO, as many object instances are not segmented yet. 

\begin{figure*}[!htbp]
\begin{center}
   \includegraphics[width=\textwidth]{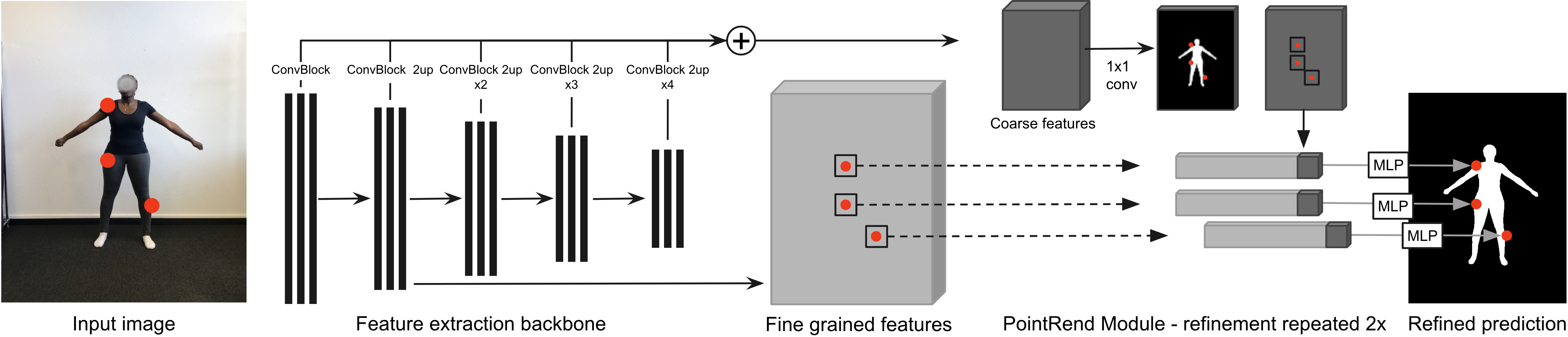}
\end{center}
   \caption{ALiSNet Architecture.  The first stage of our method is the feature extractor backbone which produces features at 5 resolution levels. These features are convolved and added together to form the coarse features. The coarse features are projected to a coarse segmentation using a $1\times1$ convolution.  Uncertain points on the coarse segmentation mask are selected and refined using PointRend.  During a PointRend refinement step, coarse predictions and finegrained features from these locations are concatenated and fed to an MLP to obtain refined segmentation predictions.  This refinement is repeated two times. In the diagram, \emph{ConvBlock} is a conv, batch-norm, ReLu sequence and \emph{2up} is a $2\text{x}$ bi-linear upsampling.}
\label{fig:arch}
\end{figure*}

\section{\uppercase{Method}}
\subsection{Model}
Our model, ALiSNet, is a version of Semantic FPN with PointRend~\cite{kirillov2020pointrend}, simplified for on-device use. We chose Semantic FPN with PointRend as a baseline because of its high accuracy in segmenting object boundaries.
In theory, other baseline methods could also be used to show the effectiveness of our approach.

\paragraph{Semantic FPN with PointRend:}
Semantic FPN first extracts features using a backbone and further process them with a Feature Pyramid Network (FPN)~\cite{lin2017feature}. Then, a coarse segmentation map is computed from the aggregated coarse features.
PointRend then samples uncertain points on the coarse segmentation and concatenates fine-grained features from the FPN with the coarse predictions at each location and uses this as an input to a classifier to refine the prediction at this location.


\paragraph{Changes to Baseline for On-device Use:}
In this work, we use three approaches to address the problem of reducing the model size while preserving segmentation accuracy.
First, we take advantage of a mobile feature extraction backbone to replace the ResNeXt101~\cite{xie2017aggregated} feature extractor in our baseline.
Second, we quantize our model using quantization-aware training, allowing us to replace 32bit floating point parameters with equivalent int8 representations. We choose to use quantization aware training as opposed to post-training quantization as it is known to lead to more accurate results.
Third, we replace the feature pyramid network used in the original model with a simpler aggregation step, skipping the FPN top-down path.
The final ALiSNet architecture is shown in~\autoref{fig:arch}.

\paragraph{Training Loss:}
Our training loss includes the segmentation loss between the predicted and ground-truth labels, and the PointRend loss. For the segmentation loss, we experimented with cross-entropy and focal loss~\cite{lin2017focal}, and found cross-entropy to be more stable during training. We therefore used the latter for further trainings. The PointRend loss is taken from the reference implementation of PointRend in detectron2 and contains the sum of cross-entropies between predicted and ground-truth labels of all points refined during the refinement process for each sample in the mini-batch.  


\subsection{Data}
\label{datasection}

An important element in making our approach successful is to pre-train on a large-scale coarsely annotated dataset and fine-tune on a high-quality specific dataset.

\paragraph{Pretraining on COCO:}
Following other segmentation methods~\cite{kirillov2020pointrend} we base our work on MS COCO. Out of all images in COCO, we only use those containing at least one person (around $60\text{k}$ images). The COCO default annotation format is designed for instance segmentation. Thus, we merge the segmentation masks of human instances in each image to create corresponding segmentation masks for semantic segmentation.
%

\begin{figure}[!htbp]
\begin{center}
   \includegraphics[width=0.4\linewidth]{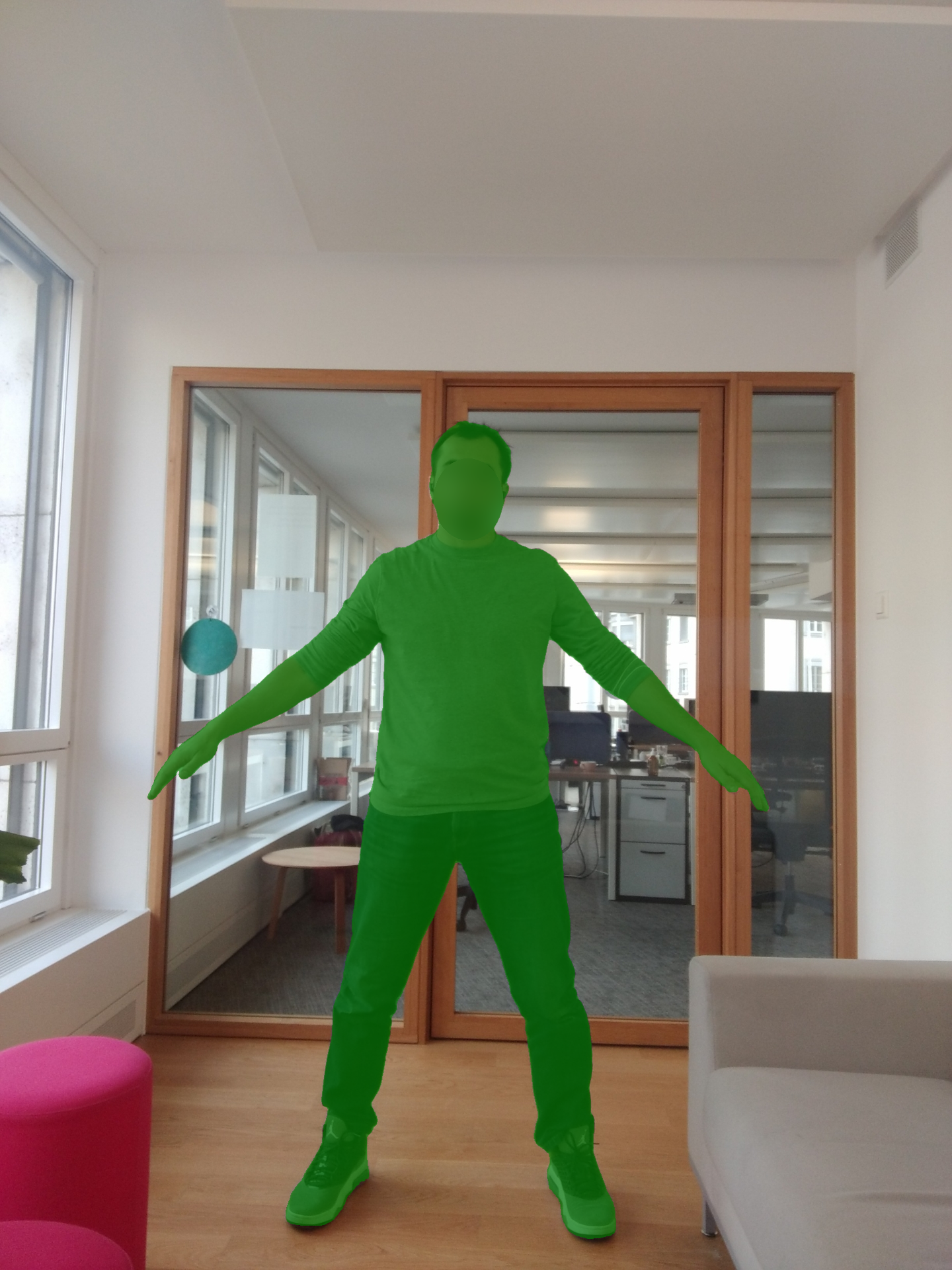}
   \includegraphics[width=0.4\linewidth]{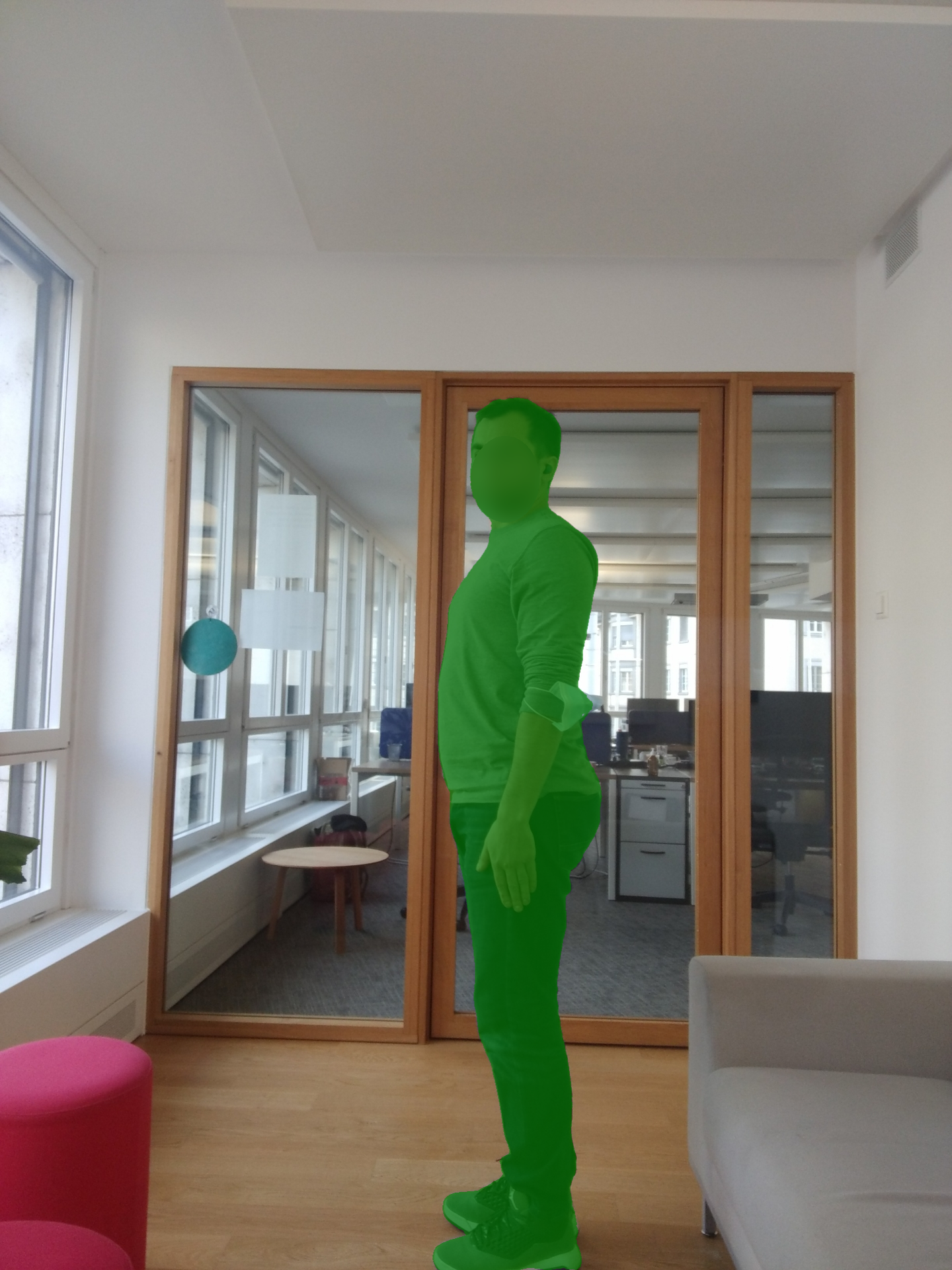}
\end{center}
\label{fig:poseexamples}
   \caption{left: Examples of front and side view images in the target poses. Ground-truth annotation masks are overlaid with green color onto the pictures.}
\label{fig:finetune-data-example}
\end{figure}

\paragraph{Small scale in-house dataset:}
We observe that even though object instance coverage in COCO is very high, there are disadvantages when using only that data for our task: Segmentation annotations are not pixel accurate since the objects are annotated using polygons. The scale of objects can be extremely diverse, ranging from objects of ~$10$ pixels in height to objects covering almost the entire image. The diversity of human poses and occlusions is extreme and most images contain only a small body part or crowds of people. This is helpful for general segmentation tasks but our experiments show that it limits the accuracy of on-device models in more controlled tasks such as body shape estimation where pose, viewing angle and scale of the body does not vary much.

To overcome these limitations of large-scale general purpose datasets, we make use of a small high-quality dataset focused on enabling accurate human body segmentation for our task. To build this complementary dataset, we use an in-house mobile app with interactive video features to guide the users to stand in the correct front and side view poses at the right distance to be fully in the frame. The app was made available for both iOS and Android, powered by real-time pose estimation models native to the OS. Images are taken with portrait (vertical) framing format. The calculated pose-keypoints are available for the captured images and can be used in downstream tasks too. To protect participants' privacy, the images were cropped around the bounding box containing the person. 
The bounding boxes are calculated from predicted pose-keypoints in the app and enlarging the box by $10\%$ margin on each side.
This dataset includes $6147$ images which are randomly distributed in train/validation/test splits with $60$/$20$/$20$\% ratios respectively. The number of front and side view images is almost balanced, and the ratio of male/female participants is $45$/$55$\%. The annotation is performed by an expert annotation team, and the quality of segmentation masks was checked by two quality expert groups. An example of this data is shown in~\autoref{fig:finetune-data-example}.

\section{\uppercase{Experimental Setup}}
\subsection{Implementation details}

For this paper, we used the detectron2 framework~\cite{wu2019detectron2} built on top of PyTorch, which includes the reference implementation of PointRend.


On the mobile side, the model is executed by the PyTorch Mobile interpreter to facilitate the deployment of developed models. To that end, the model is first converted by the TorchScript compiler and the resulting model graph is loaded by the mobile interpreter. Because the iterative PointRend head contains control flows based on the input, we have to use scripting instead of tracing for computation graph generation. Scripted models are not fully optimized for runtime, therefore the performance sometimes is lower than traced models. For the quantization of the model we use the QNNPACK~\cite{qnnpack} backend which is optimized for ARM CPUs available in mobile devices.

 \begin{table*}[!htbp]
 \begin{center}
 \begin{tabular}{|l|c|c|}
 \hline
 Model configuration & Size (MB) & mIoU \\
 \hline\hline
ResNeXt101 + FPN-SemSeg + PointRend & 351.7 & 97.8 $\pm$ 1.0 \\
Replace ResNeXt101 with MnasNet-B1 & 52.0 & 97.7 $\pm$ 0.9 \\
Quantization-Aware-Training & 12.9 & 97.7 $\pm$ 0.9 \\
Remove FPN-top-down path (ALiSNet) & \textbf{4.0} & 97.6 $\pm$ 1.0 \\
  \hline
 Google MLKit (BlazePose-Segmentation)-accurate & 27.7 & 93.9 $\pm$ 5.3 \\
 Google MLKit (BlazePose-Segmentation)-balanced & 6.4 & 93.7 $\pm$ 5.9 \\
 Apple Person Segmentation (accurate) & - & 94.3 $\pm$ 5.9 \\
 Apple Person Segmentation (balanced) & - & 94.4 $\pm$ 5.7 \\
 \hline
 \end{tabular}
 \end{center}
    \caption{Effect of each model change step on the accuracy and size of model. mIoU values are reported as mean $\pm$ std in percent.}
 \label{tab:model_design}
 \end{table*}

 \begin{table*}[!htbp]
 \begin{center}
 \begin{tabular}{|l|c|c|c|}
 \hline
 Model &  Size & mIoU      & mIoU           \\
       &  (MB) & with COCO & with fine-tuning\\
 \hline\hline
 ResNeXt101 + FPN-SemSeg + PointRend & 351.7 & 94.0 $\pm$ 3.8 & 97.8 $\pm$ 1.0 \\
  MobileNetV3 + FPN-SemSeg + PointRend & 35.1 & 91.2 $\pm$ 6.2 & 97.7 $\pm$ 1.1 \\
 (Q) MnasNet-B1 + SemSeg + PointRend (ALiSNet) & \textbf{4.0} & 90.0 $\pm$ 6.8 & 97.6 $\pm$ 0.9 \\
 \hline
 \end{tabular}
 \end{center}
  \caption{Effect of fine-tuning on our dataset. First mIoU  column: results after training on COCO only. second mIoU column: result after fine-tuning on our dataset. (Q) indicates the model is quantized using quantization-aware-training.}
 \label{tab:fine-tuning}
 \end{table*}

\subsection{Model Training}
As described in~\autoref{datasection}, we augment training our models on COCO with fine-tuning on our in-house dataset. We also experiment with CNN backbones that are pre-trained on ImageNet~\cite{deng2009imagenet} classification tasks. All experiments are done in machines in Amazon AWS with $4$ V100 GPUs totalling $128$ GB graphic memory. The mini-batch size of training is set to $8$ for large models (e.g. ResNeXt101 backbone) and $16$ for smaller backbones (i.e. MnasNet, MobileNet, FBNet). The base learning rate is set to $0.01$ for the case of batch-size = $8$ and $0.02$ for batch-size = $16$ following the recommendation of a linearly scaling learning rate~\cite{DBLP:journals/corr/GoyalDGNWKTJH17}. During training we augment the data using default augmentation tools provided by detectron2. This includes random resizing, horizontal flip, color jitter and brightness and saturation change. The range of sizes for the shorter side of image in resizing is set randomly from a predefined list (between $120$ and $800$) while keeping the longer side under $1024$ and the scale-factor between $0.5$ and $4$. This prevents too much down-sampling of our high-resolution images during training.

For the fine-tuning step, as a standard practice we experimented with freezing the first $N$ $(N\leq2)$ stages of the backbone to improve generalization of the model and avoid over-fitting but we have observed that not freezing any layer can improve the generalization results in our case. We also experimented with reducing the learning-rate by $\times10$ for the fine-tuning step compared to the pre-training learning rate. However, we found that the model converged quicker and to better results when we did not reduce the learning rate.

 \subsection{Evaluation}

 We evaluate our models with mean Intersection Over Union (mIoU) which is defined
 in~\autoref{eq:iou_formula}. This metric is in the $[0,1]$ range and then reported as percentage.

 \begin{equation}
 \label{eq:iou_formula}
     mIoU = \frac{1}{N}\sum_{i=1}^{N}\frac{|pred_i \cap GT_i|}{|pred_i \cup GT_i|} \times 100
 \end{equation}
 where $pred$ and $GT$ are the prediction and ground-truth segmentation masks of sample $i$ respectively. During evaluation, images are sized to $1024$ on height after cropping them to the person bounding box. 

\section{\uppercase{Experimental Results}}

In this section, first we explore the effect of different aspects of our model. Then we do a quantitative and qualitative comparison of our method with two other related person-segmentation methods.

\subsection{Effect of Model Design Choices}
\paragraph{Reduction of Size of Components:}
As shown in~\autoref{tab:model_design}, starting from the baseline model ($351.7$MB), we first obtain a model size of $52.0\text{MB}$ by replacing the ResNeXt101 feature backbone with MnasNet, saving around $300$ MB. Then we applied Quantization Aware Training, which further shrinks the model size by $\times$4, resulting in $12.9\text{MB}$, as we replace $32$ bit floating point with int8 representation.

We then show that the top-down branch of FPN which combines high-level semantic features with low-level features can be removed from the model with only $0.1\%$ reduction in accuracy. We argue that in our model, PointRend carries the job of merging high-level and low-level features, thus making the top-down path of FPN mostly redundant. Furthermore, scale of persons in our data does not vary enough to require FPN-top-down features.


\paragraph{Fine-tuning:}
We first train all the models on the COCO person class. In~\autoref{tab:fine-tuning}, we show the effect of fine-tuning these models on our high-quality task-specific dataset. It is clear that the fine-tuning significantly improves the mIoU, and the effect is greater for smaller models. 

\paragraph{PointRend:}
Although the PointRend module adds to the  $~0.2\text{MB}$ size and  $20\%$ to the runtime of our method, we use it, as \autoref{tab:pointrend-effect} shows that it adds around $0.3\%$ mIoU to the model accuracy.

\begin{table}[!htbp]
\begin{center}
\begin{tabular}{|l|c|c|}
\hline
Backbone & FBNetV3 & MnasNet-B1 \\
\hline\hline
no PR & 97.4 $\pm$ 1.1 & 97.4 $\pm$ 1.0 \\
With PR & 97.7 $\pm$ 0.9 & 97.7 $\pm$ 0.9 \\
\hline
\end{tabular}
\end{center}
\caption{Effect of PointRend on accuracy of Semantic FPN without and with PointRend. Both models are quantized.}
\label{tab:pointrend-effect}
\end{table}



\paragraph{CNN backbones choice:}

We compared the effect using MobileNetV3~\cite{howard2019searching}, MnasNet~\cite{tan2019mnasnet}, FBNetV3~\cite{dai2021fbnetv3} as mobile-friendly backbone feature extractors. \autoref{tab:backbones} shows that all of the mobile feature extractors have around the same performance in our task, which is only $0.1\%$ lower than the much larger ResNeXt101.  We chose to use MnasNet due to lower variance and its availability in the the torchvision framework.

\begin{table}[hbtp!]
\begin{center}
\begin{tabular}{|l|c|}
\hline
Model backbone & mIoU \\
\hline\hline
ResNeXt101  & \textbf{97.8} $\pm$ 1.0 \\
MobileNetV3  & 97.7 $\pm$ 1.1 \\
(Q) FBNetV3  & 97.7 $\pm$ 0.9 \\
(Q) MnasNet-B1 & 97.7 $\pm$ 0.9 \\
\hline
\end{tabular}
\end{center}
  \caption{Effect of different CNN backbones on accuracy of the Semantic FPN segmentation model.}
\label{tab:backbones}
\end{table}
 
\subsection{Runtime on mobile devices}
\begin{figure}[!htbp]
\begin{center}
\includegraphics[width=\linewidth]{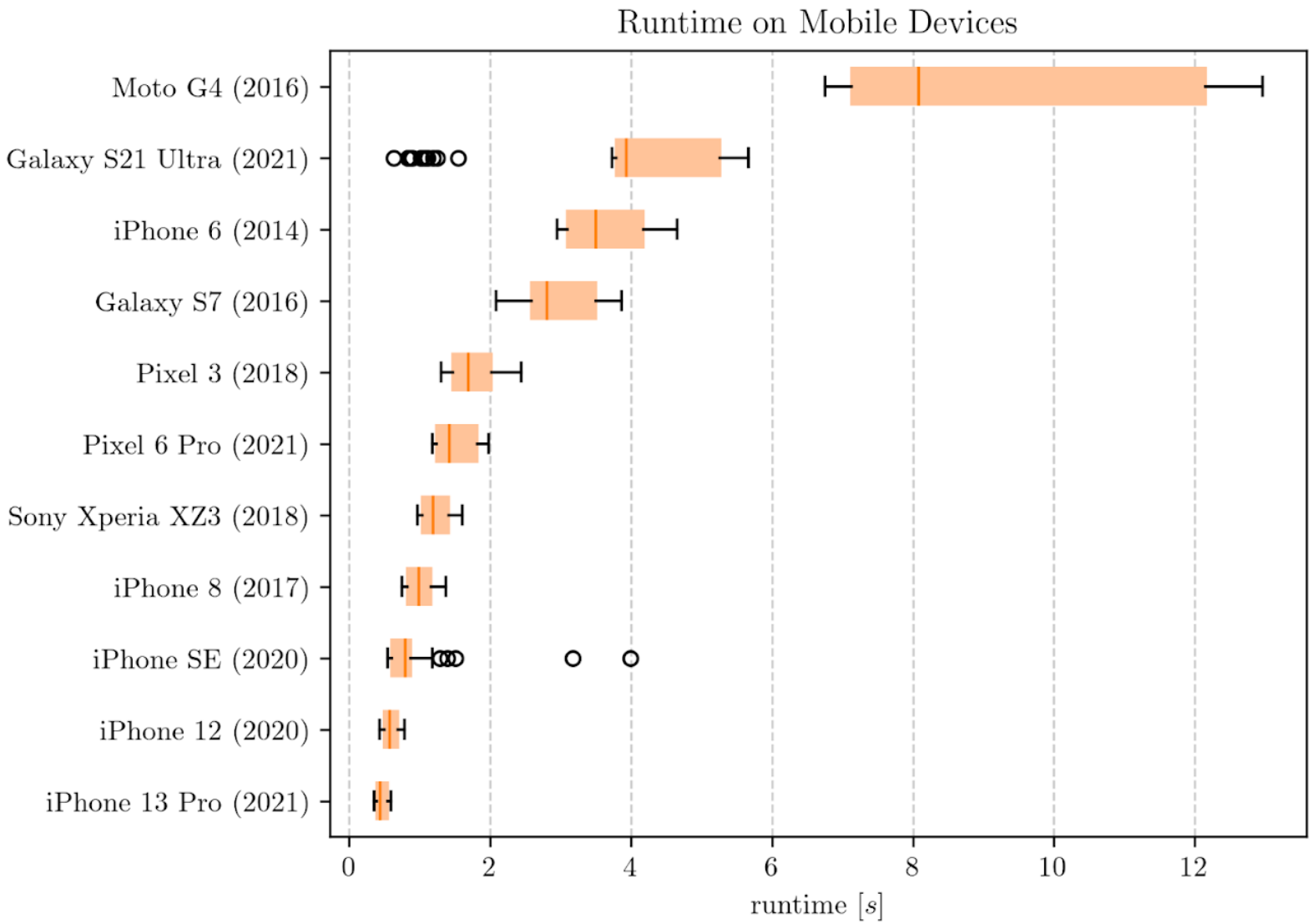}
\end{center}
\caption{Runtime on mobile devices. Our method runs in less than two seconds on most modern devices. As of October 2022, iPhone 13 is the latest iPhone available on the Amazon Device Farm.}
\label{fig:runtime}
\end{figure}

We evaluated our model on a set of real mobile devices provided by the AWS Device Farm\footnote{https://aws.amazon.com/device-farm/}. 
The distribution of runtime is shown in~\autoref{fig:runtime}. For this evaluation, 90 high-resolution images from the dataset are processed using our in-house evaluation mobile app. Images are resized to $2\text{k}$ resolution in height while preserving the aspect ratio and then cropped to the person bounding box. The cropped images are passed to the model, where they are resized to $1024$ in height internally before segmentation. As the bounding box of the person varies between images, a significant variance of the runtime on a single device can be noticed. On recent iPhones the model runs well below $1\text{s}$ and for an older Android phone like the Moto G4 the mean is around $8\text{s}$. In iPhone SE and Galaxy S21 there are some outliers in runtime, for reasons we were not able to determine. The model is running on CPU mode due to limited support of PyTorch for mobile-GPU in quantized models.

\begin{figure}[!htbp]
\centering
\begin{tabular}{ccc}
\includegraphics[height=3.29cm]{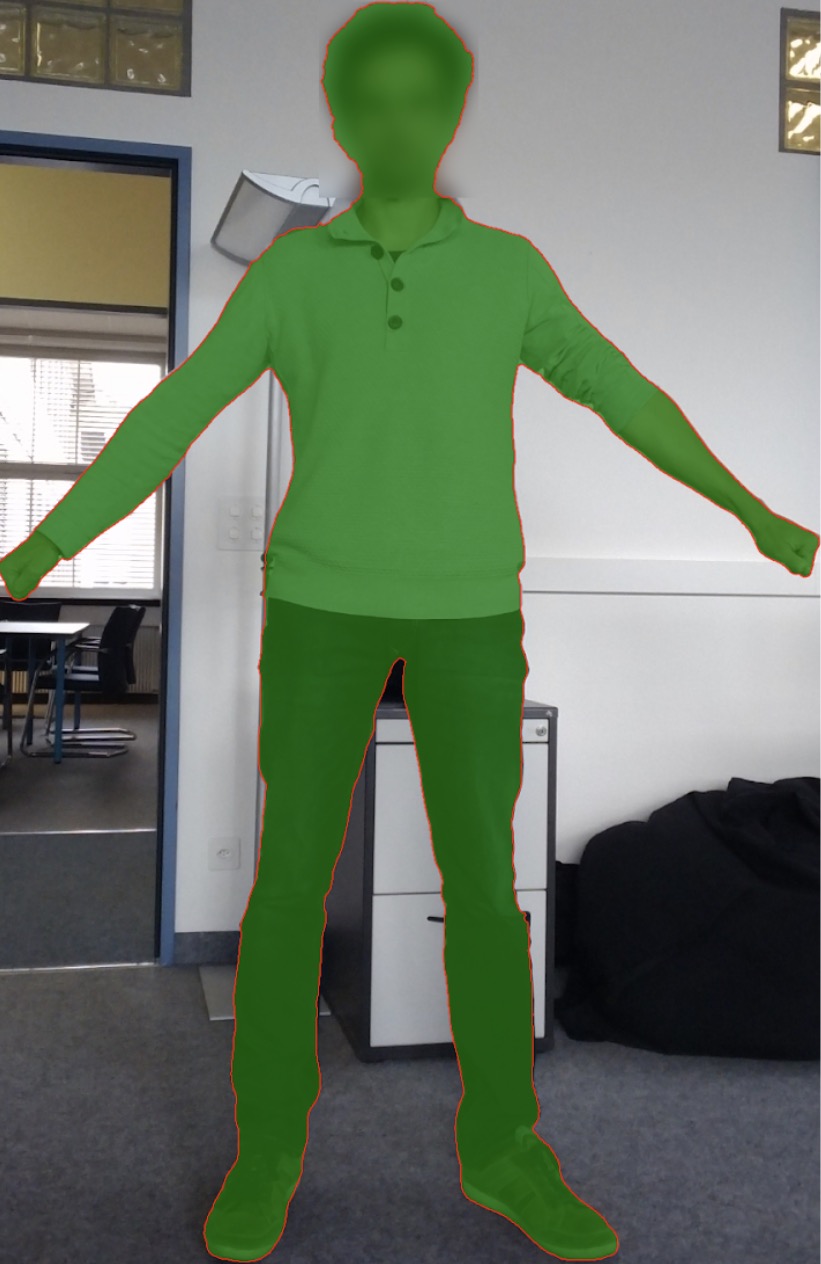} &
\includegraphics[height=3.29cm]{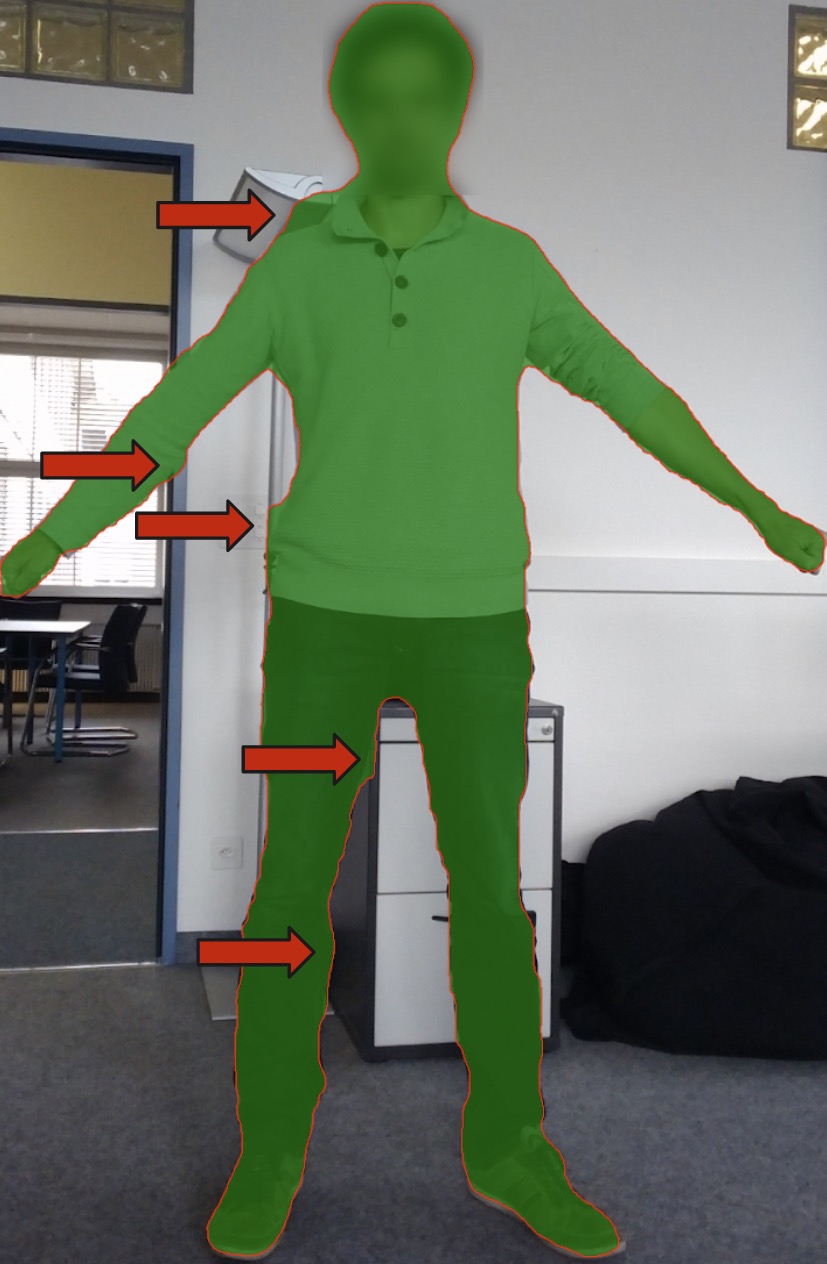} &
\includegraphics[height=3.29cm]{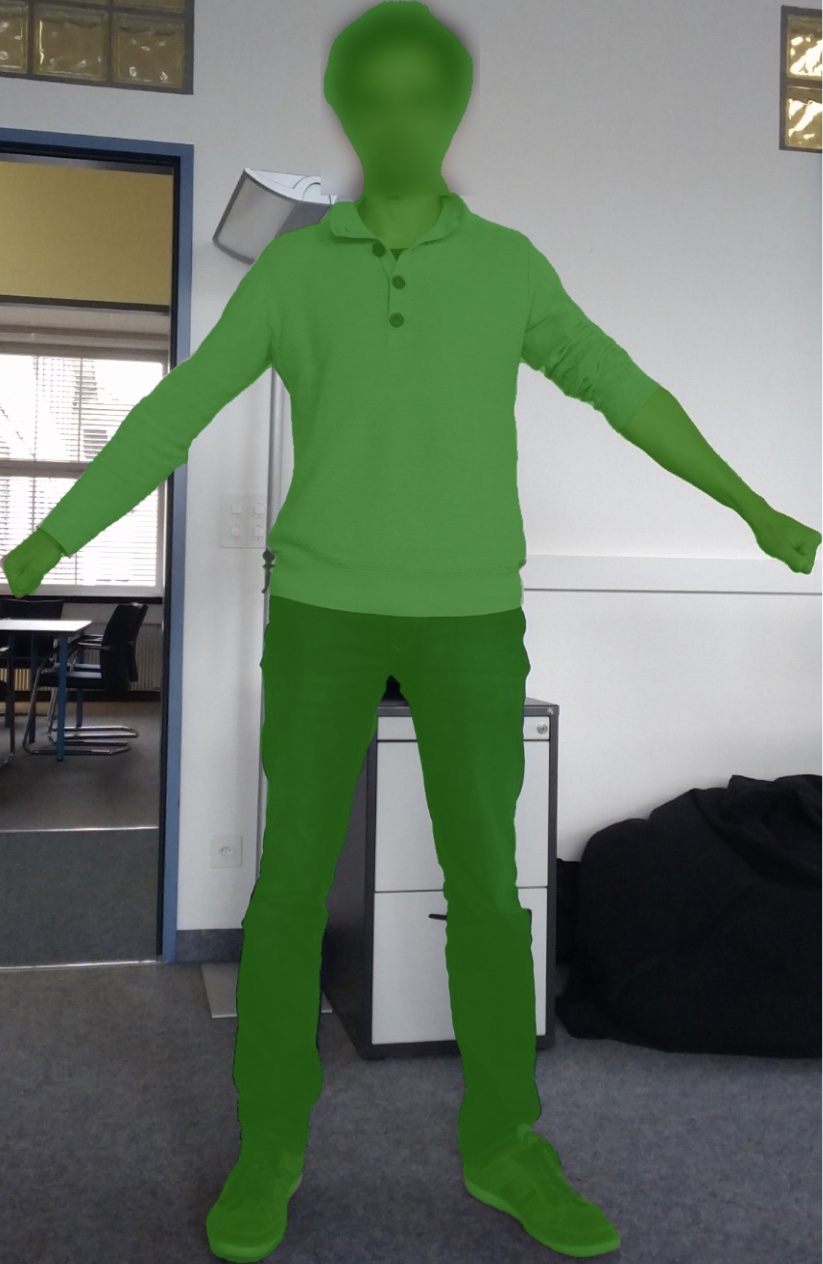} \\
\includegraphics[height=3.5cm]{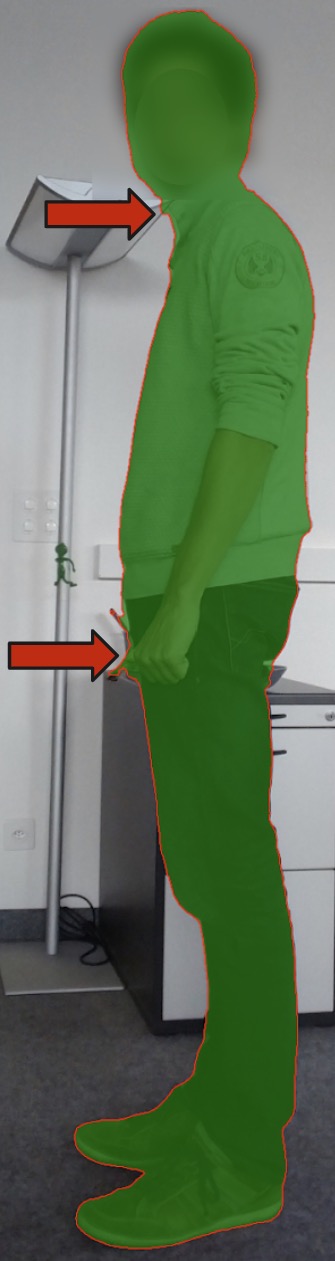} &
\includegraphics[height=3.5cm]{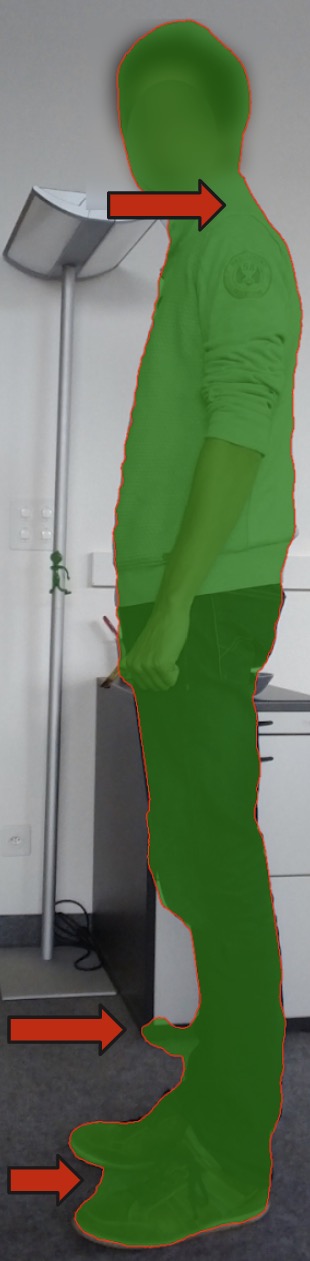} &
\includegraphics[height=3.5cm]{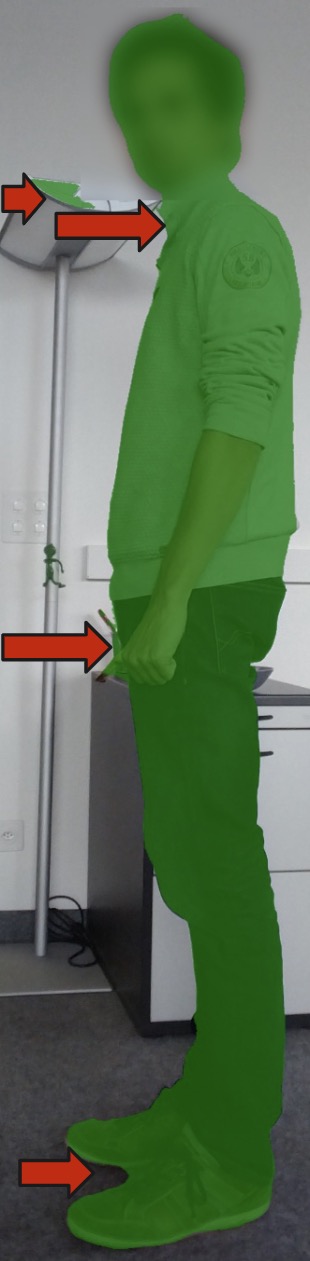} \\
ours & BlazePose & Apple

\end{tabular}
   \caption{Comparison of segmentation methods. Red arrows indicate mis-segmented regions. In these images, the threshold for the Apple segmentation algorithm (balanced) was set to 0.6 to obtain the best results. In the quantitative evaluation, the value is 0.3}
   \label{fig:compare-with-mlkit}
\end{figure}

\begin{figure*}[!htbp]
\centering

\begin{tabular}{ccc|ccc|ccc|ccc}
\includegraphics[height=2.1cm]{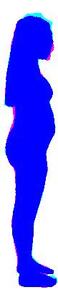}
 & 
\includegraphics[height=2.1cm]{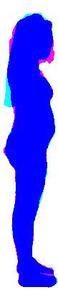}
 & 
\includegraphics[height=2.1cm]{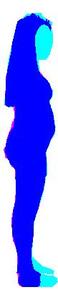}
&
\includegraphics[height=2.1cm]{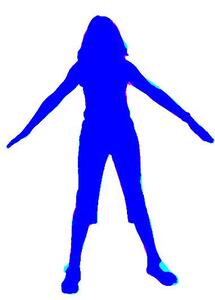}
 & 
\includegraphics[height=2.1cm]{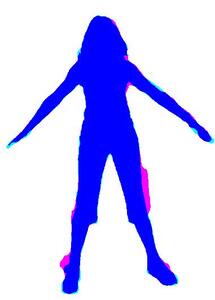}
 & 
\includegraphics[height=2.1cm]{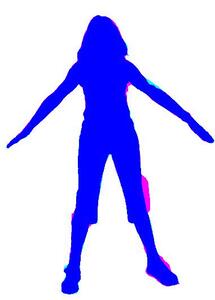}
&
\includegraphics[height=2.1cm]{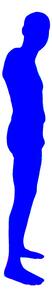}
 & 
\includegraphics[height=2.1cm]{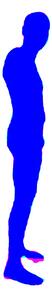}
 & 
\includegraphics[height=2.1cm]{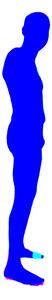} & 
\includegraphics[height=2.1cm]{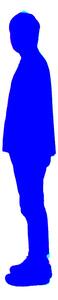}
 & 
\includegraphics[height=2.1cm]{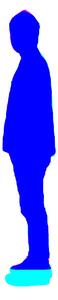}
 & 
\includegraphics[height=2.1cm]{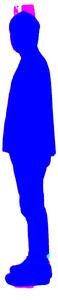}
\\
ours & BP & Apple & 
ours & BP & Apple &
ours & BP & Apple &
ours & BP & Apple \\

95.7 & 93.8 & 87.3 &
96.4 & 93.2 & 95.5 & 
98.6 & 96.7 & 97.6 & 
98.8 & 90.9 & 97.0

\end{tabular}
    \caption{
    Overlays of \textcolor{pink}{predicted segmentations} over the \textcolor{blue}{ground truth annotations}, \textcolor{purple}{blue intersection}, from our in-house dataset, from our method, BlazePose (BP) and Apple, with the IoU.  We ranked the images in the test set by their their mIoUs using AlisNet, and displayed the 5th, 10th, 90th and 95th-\% images in this ranking.  As the photos are confidential, we show only the silhouettes here. After data collection, all images were anonymized using face-blur, which is seen in the Apple Segmentation in the first image. In one of the BP segmentations you can see the issue of bounding box prediction cutting out part of the feet.}
    \label{fig:silhouette_overlays}
\end{figure*}

\subsection{Comparison to BlazePose and Apple Person Segmentation}

We compare with two on-device person segmentation methods.

BlazePose~\cite{blazepose} is a on-device real-time body pose tracking method, which provides a segmentation prediction option. We compare two settings of the BlazePose model, balanced and accurate, which have model sizes of $6.4\text{MB}$ and $27.7\text{MB}$ respectively.

We also compare to Apple Person Segmentation which was made available in iOS15. Information about the details of this model is not available.  

The output segmentation maps of these methods are probabilistic, and need to be thresholded to compute the final binary silhouette maps. Thresholds were determined using a sweep of values and were set to $0.5$ for BlazePose and $0.3$ for Apple.

It is not possible to fine-tune either of these methods on our data.

As seen in~\autoref{tab:model_design}, both BlazePose and Apple Segmentations have a much lower mIoU than AliSNet on our dataset, while having a much higher standard deviation.  This indicates that fine-tuning on our data allows our model to avoid certain mistakes in segmentation. Neither of the compared methods are designed for the use-case of segmentation for accurate human body measurement. BlazePose is optimized for real-time pose estimation, which means it lies on a different point of the performance-accuracy trade-off. Apple Person Segmentation is designed to power Portrait Mode. 
Segmentation examples from all three methods are shown  in~\autoref{fig:compare-with-mlkit}. In the front view, both our method and Apple produce more accurate segmentations than BlazePose.
In the side view, we see that all three methods have difficulty with background objects, and that the Apple method produces artifacts in the lamp.

\autoref{fig:silhouette_overlays} shows overlays from the test set of our dataset. In this figure we show both ``bad" and ``good" segmentations from our model, however, we see that our segmentation has a higher IoU than the other methods, as we have trained our model on this dataset.

\begin{figure}[!htbp]
\centering
\begin{tabular}{ccc}
\includegraphics[trim={30cm 30cm 30cm 30cm},clip,height=3.6cm]{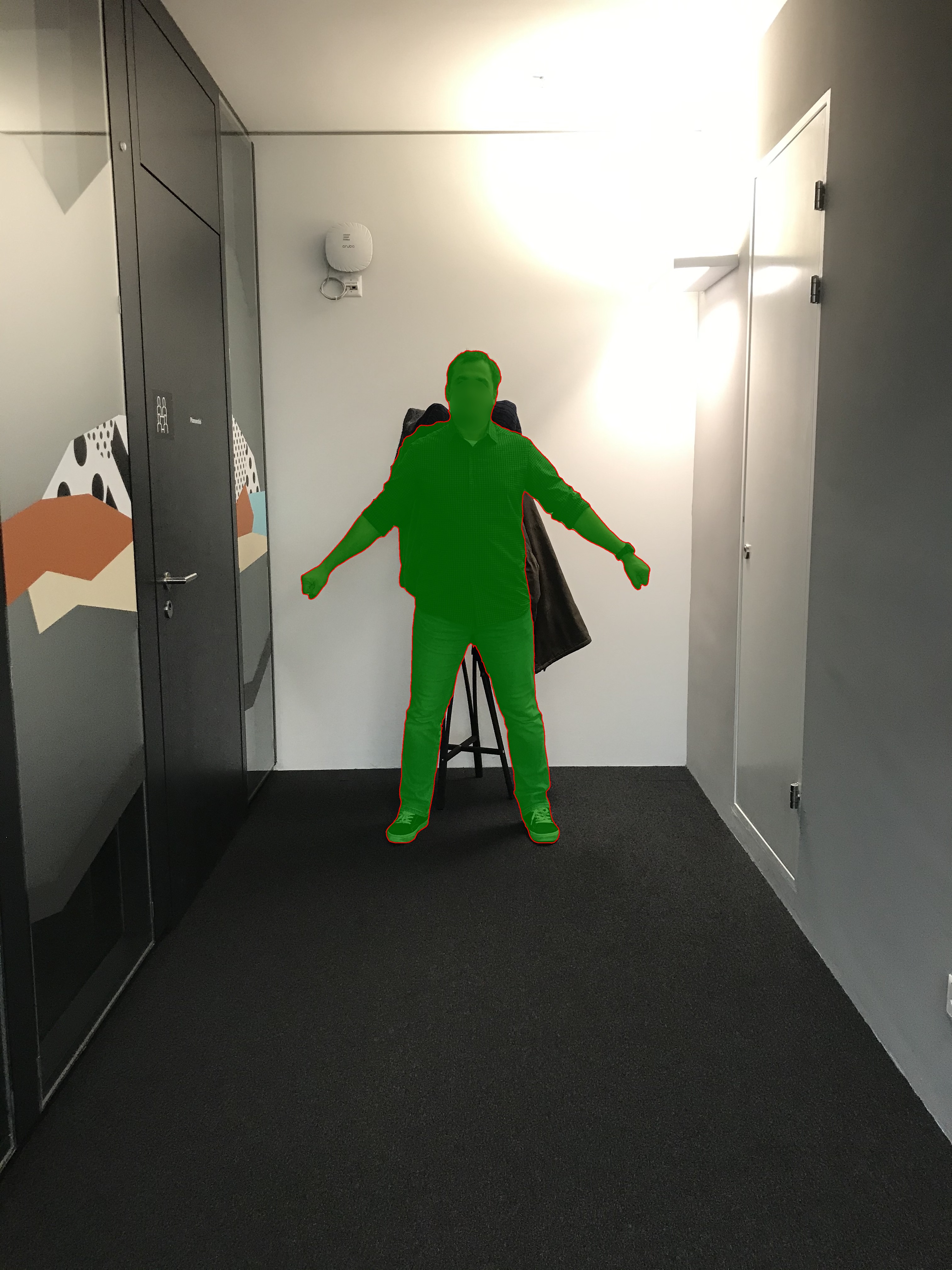} &
\includegraphics[trim={30cm 30cm 30cm 30cm},clip,height=3.6cm]{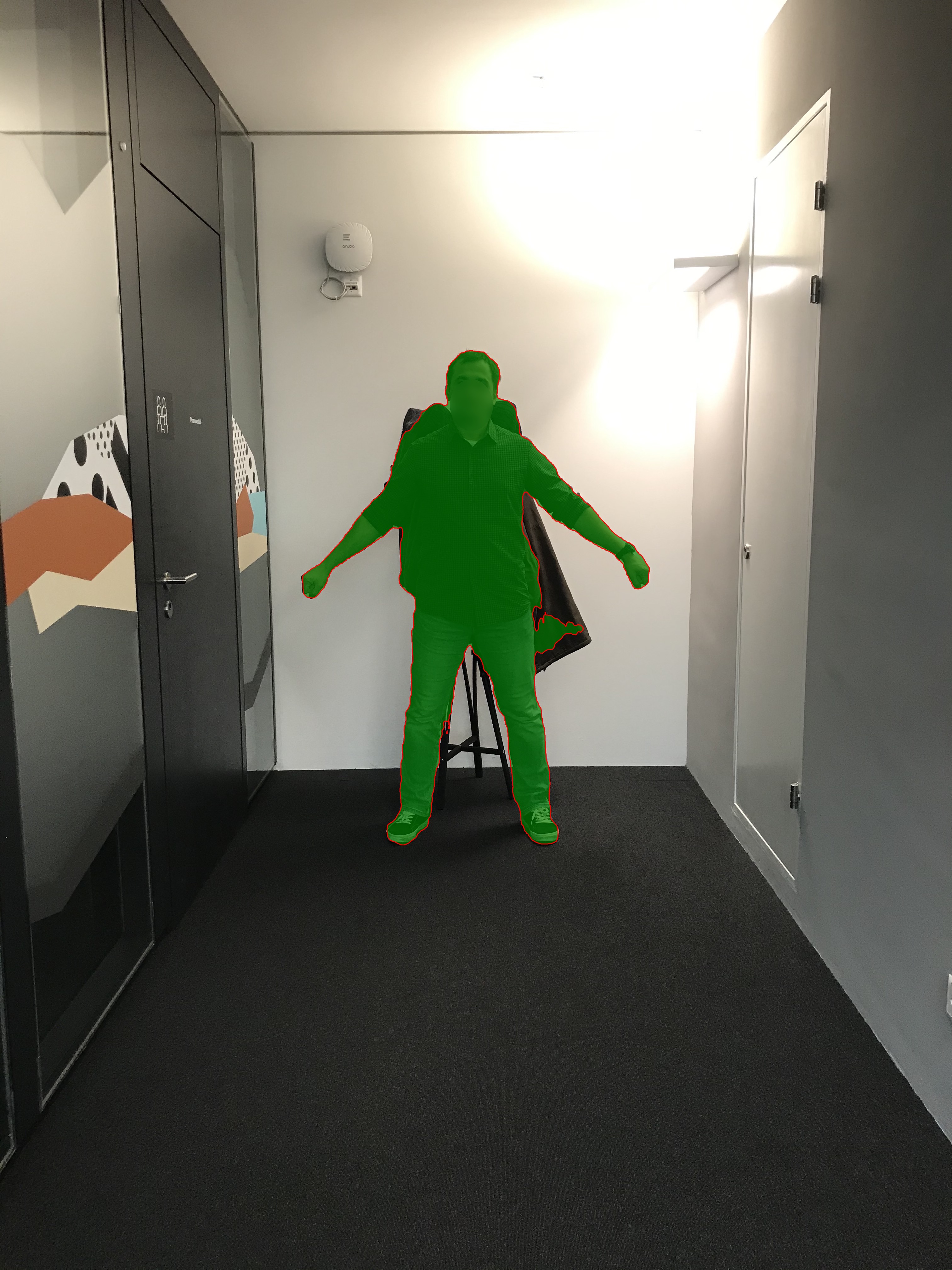} &
\includegraphics[trim={30cm 30cm 30cm 30cm},clip,height=3.6cm]{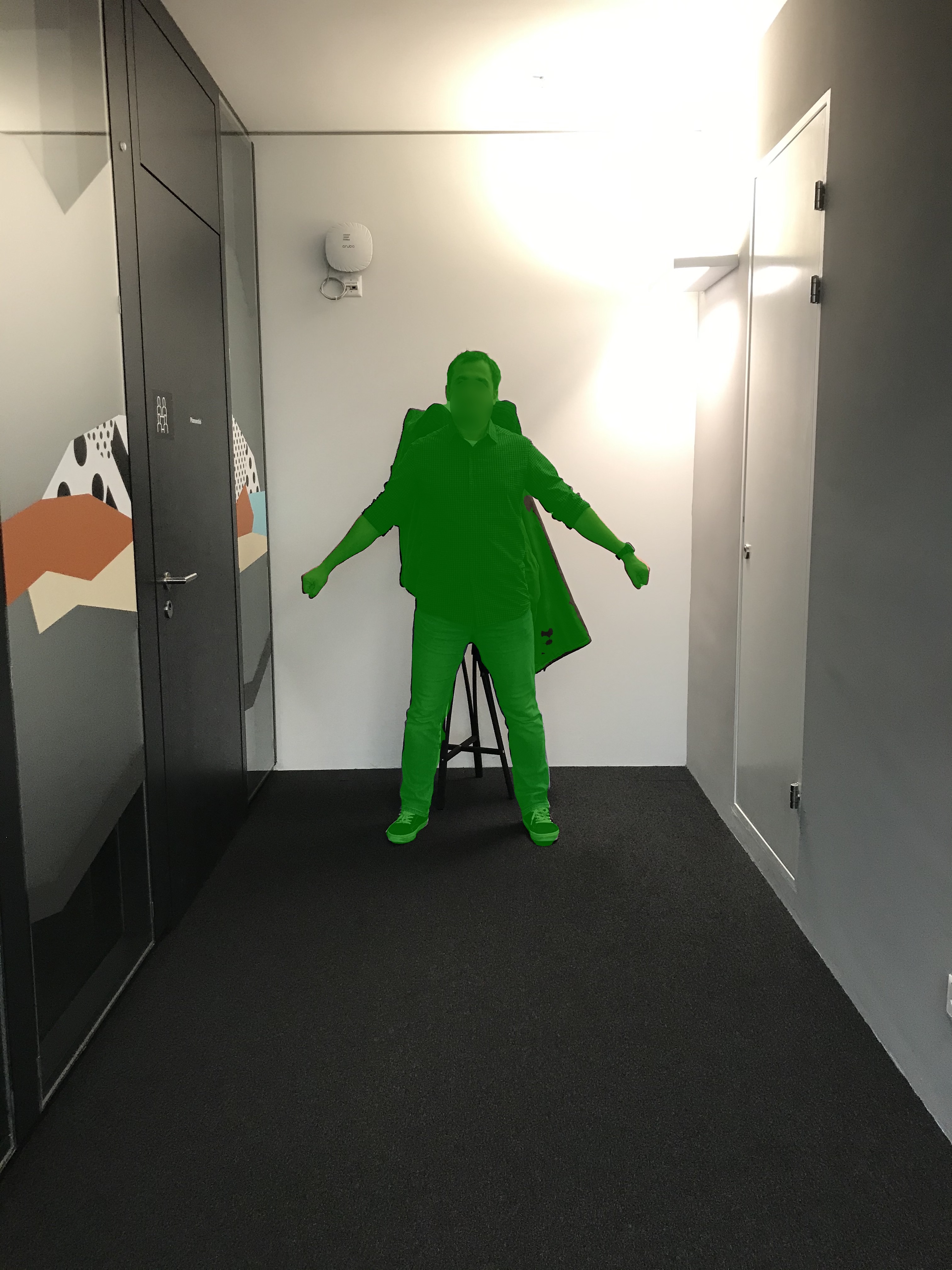} \\
\includegraphics[trim={20cm 10cm 10cm 10cm},clip,height=3.6cm]{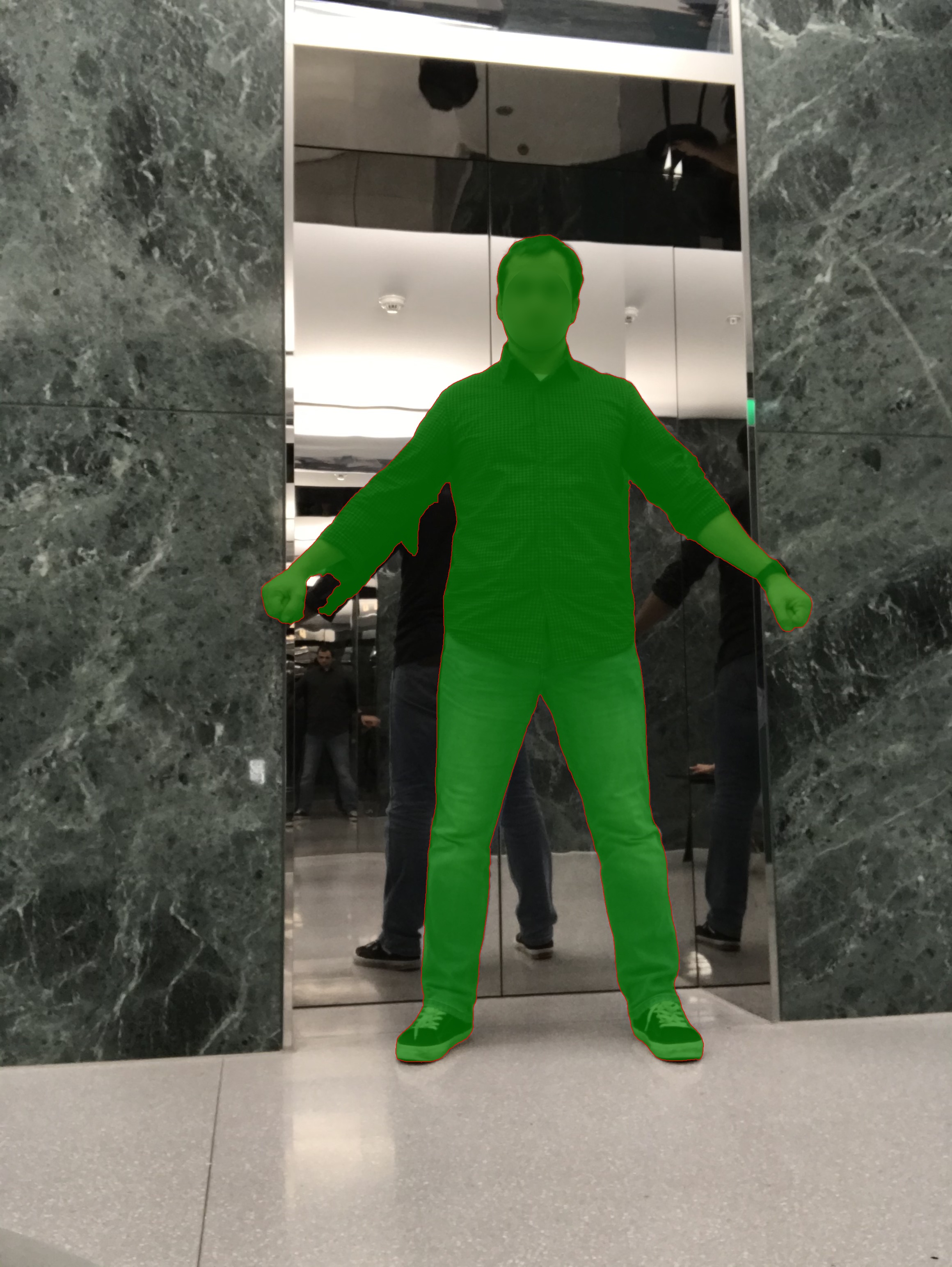} &
\includegraphics[trim={20cm 10cm 10cm 10cm},clip,height=3.6cm]{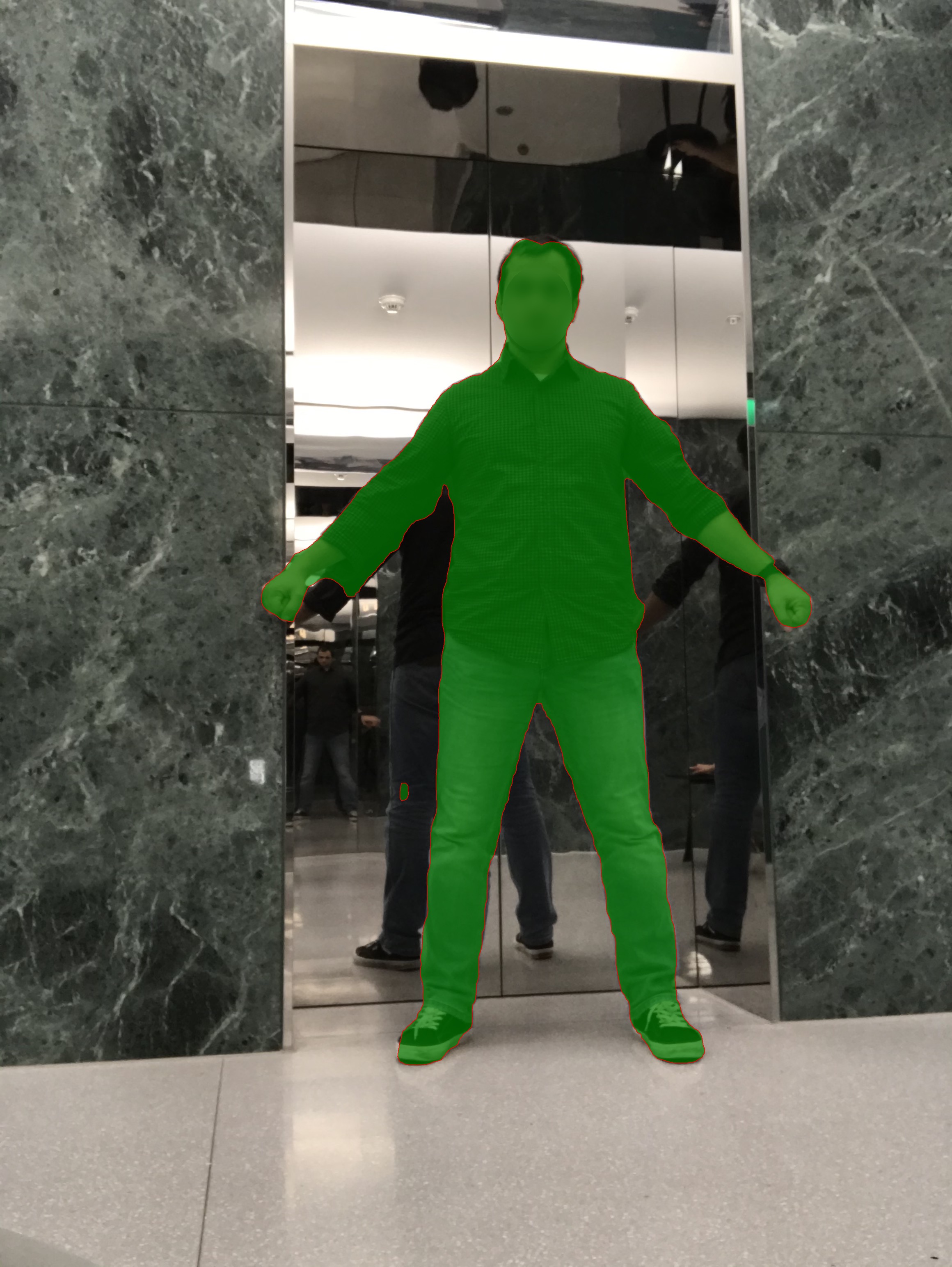} &
\includegraphics[trim={20cm 10cm 10cm 10cm},clip,height=3.6cm]{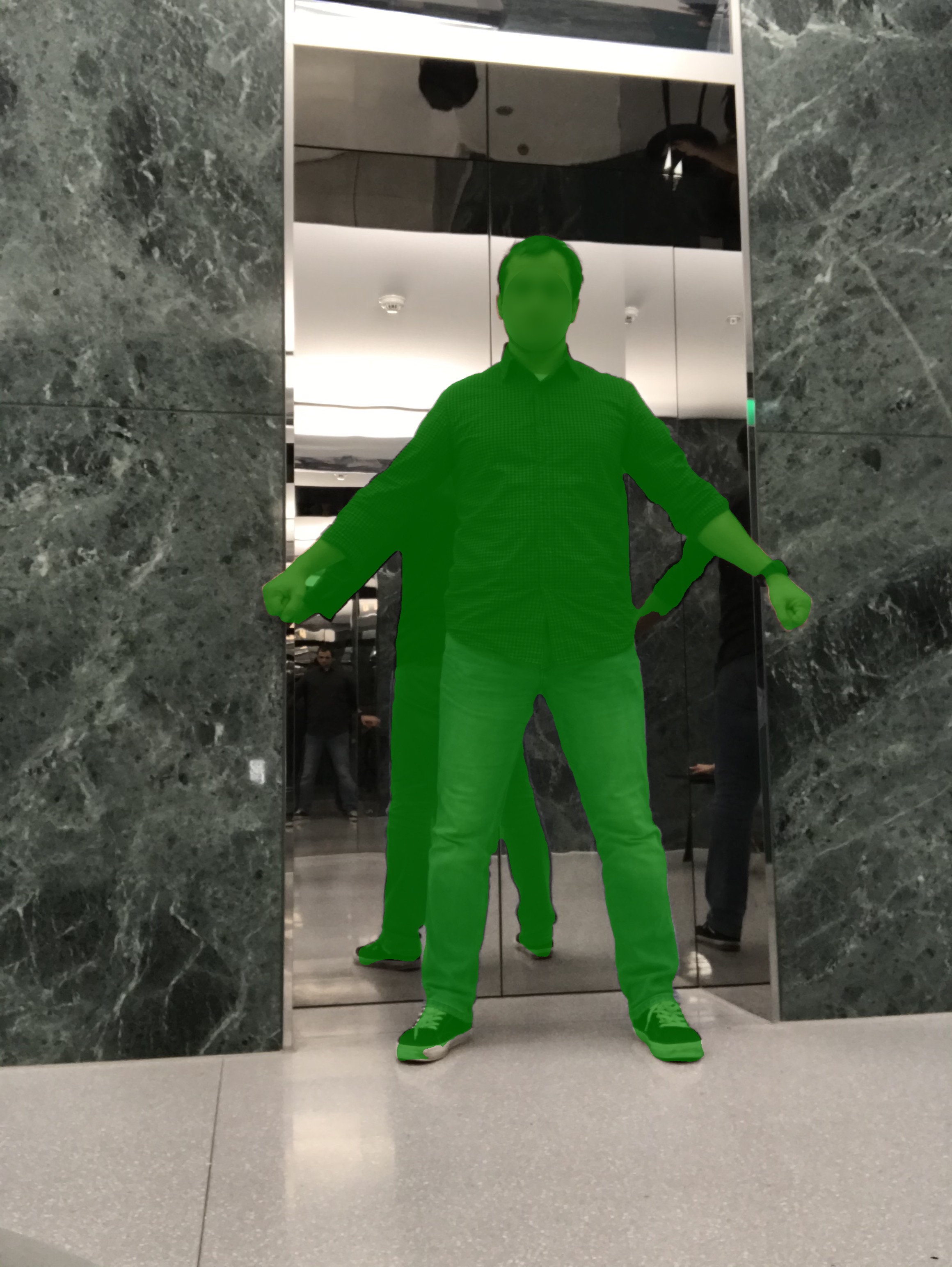} \\
ours & BlazePose & Apple\\

\end{tabular}
   \caption{Examples of failure cases for segmentation algorithms. In our user collected dataset, people take pictures at home, and sometimes have clothes (top) or mirrors (bottom) in the background, which cause the segmentation method to not work correctly.  This is a shortcoming of all the methods we evaluate here.}
   \label{fig:failures}
\end{figure}

\section{\uppercase{Discussion}}

We presented a method for accurate mobile human segmentation along with a set of general steps that can be used to simplify existing large-scale models for on-device applications. 

Although our model handles most images well, there are cases with confusing background textures where our method and other methods fail, as shown in~\autoref{fig:failures}. Other challenging conditions include dim lighting, dark shadows or other image distortions.
Improving the performance under these conditions would be an important future direction.

In the future we will be experimenting with on-device segmentation models for accurate body shape and measurement estimation. 

\section*{\uppercase{Acknowledgements}}
We would like to thank Julia Friberg for her contributions in evaluation of the models on real mobile phones.

\bibliographystyle{apalike}
{\small
\bibliography{ALiSNet}}

\begin{thebibliography}{}

\bibitem[Apple, 2022]{apple_segmentation}
Apple (2022).
\newblock {Applying Matte Effects to People in Images and Video}.
\newblock
  \url{https://developer.apple.com/documentation/vision/applying_matte_effects_to_people_in_images_and_video}.

\bibitem[Bazarevsky et~al., 2020]{blazepose}
Bazarevsky, V., Grishchenko, I., Raveendran, K., Zhu, T., Zhang, F., and
  Grundmann, M. (2020).
\newblock Blazepose: On-device real-time body pose tracking.
\newblock {\em CoRR}, abs/2006.10204.

\bibitem[Dai et~al., 2021]{dai2021fbnetv3}
Dai, X., Wan, A., Zhang, P., Wu, B., He, Z., Wei, Z., Chen, K., Tian, Y., Yu,
  M., Vajda, P., et~al. (2021).
\newblock Fbnetv3: Joint architecture-recipe search using predictor
  pretraining.
\newblock In {\em Proceedings of the IEEE/CVF Conference on Computer Vision and
  Pattern Recognition}, pages 16276--16285.

\bibitem[Deng et~al., 2009]{deng2009imagenet}
Deng, J., Dong, W., Socher, R., Li, L.-J., Li, K., and Fei-Fei, L. (2009).
\newblock Imagenet: A large-scale hierarchical image database.
\newblock In {\em 2009 IEEE conference on computer vision and pattern
  recognition}, pages 248--255. Ieee.

\bibitem[Dibra et~al., 2016]{dibra2016hs}
Dibra, E., Jain, H., {\"O}ztireli, C., Ziegler, R., and Gross, M. (2016).
\newblock Hs-nets: Estimating human body shape from silhouettes with
  convolutional neural networks.
\newblock In {\em 2016 fourth international conference on 3D vision (3DV)},
  pages 108--117. IEEE.

\bibitem[Dibra et~al., 2017]{dibra2017human}
Dibra, E., Jain, H., Oztireli, C., Ziegler, R., and Gross, M. (2017).
\newblock Human shape from silhouettes using generative hks descriptors and
  cross-modal neural networks.
\newblock In {\em Proceedings of the IEEE conference on computer vision and
  pattern recognition}, pages 4826--4836.

\bibitem[Dukhan et~al., 2020]{qnnpack}
Dukhan, M., Wu, Y., and Lu, H. (2020).
\newblock {Qnnpack: open source library for optimized mobile deep learning}.
\newblock \url{https://github.com/pytorch/QNNPACK}.

\bibitem[Goyal et~al., 2017]{DBLP:journals/corr/GoyalDGNWKTJH17}
Goyal, P., Doll{\'{a}}r, P., Girshick, R.~B., Noordhuis, P., Wesolowski, L.,
  Kyrola, A., Tulloch, A., Jia, Y., and He, K. (2017).
\newblock Accurate, large minibatch {SGD:} training imagenet in 1 hour.
\newblock {\em CoRR}, abs/1706.02677.

\bibitem[Gupta et~al., 2019]{lvis}
Gupta, A., Doll{\'{a}}r, P., and Girshick, R.~B. (2019).
\newblock {LVIS:} {A} dataset for large vocabulary instance segmentation.
\newblock {\em CoRR}, abs/1908.03195.

\bibitem[Han et~al., 2020]{han2020ghostnet}
Han, K., Wang, Y., Tian, Q., Guo, J., Xu, C., and Xu, C. (2020).
\newblock Ghostnet: More features from cheap operations.
\newblock In {\em Proceedings of the IEEE/CVF conference on computer vision and
  pattern recognition}, pages 1580--1589.

\bibitem[Howard et~al., 2019]{howard2019searching}
Howard, A., Sandler, M., Chu, G., Chen, L.-C., Chen, B., Tan, M., Wang, W.,
  Zhu, Y., Pang, R., Vasudevan, V., et~al. (2019).
\newblock Searching for mobilenetv3.
\newblock In {\em Proceedings of the IEEE/CVF International Conference on
  Computer Vision}, pages 1314--1324.

\bibitem[Ji et~al., 2019]{ji2019human}
Ji, Z., Qi, X., Wang, Y., Xu, G., Du, P., Wu, X., and Wu, Q. (2019).
\newblock Human body shape reconstruction from binary silhouette images.
\newblock {\em Computer Aided Geometric Design}, 71:231--243.

\bibitem[Kirillov et~al., 2019]{Kirillov_2019_CVPR}
Kirillov, A., Girshick, R., He, K., and Dollar, P. (2019).
\newblock Panoptic feature pyramid networks.
\newblock In {\em Proceedings of the IEEE/CVF Conference on Computer Vision and
  Pattern Recognition (CVPR)}.

\bibitem[Kirillov et~al., 2020]{kirillov2020pointrend}
Kirillov, A., Wu, Y., He, K., and Girshick, R. (2020).
\newblock Pointrend: Image segmentation as rendering.
\newblock In {\em Proceedings of the IEEE/CVF conference on computer vision and
  pattern recognition}, pages 9799--9808.

\bibitem[Knapp, 2021]{knapp2021real}
Knapp, J. (2021).
\newblock {\em Real-Time Person Segmentation on Mobile Phones}.
\newblock PhD thesis, Wien.

\bibitem[Kuznetsova et~al., 2018]{google-open-images}
Kuznetsova, A., Rom, H., Alldrin, N., Uijlings, J. R.~R., Krasin, I.,
  Pont{-}Tuset, J., Kamali, S., Popov, S., Malloci, M., Duerig, T., and
  Ferrari, V. (2018).
\newblock The open images dataset {V4:} unified image classification, object
  detection, and visual relationship detection at scale.
\newblock {\em CoRR}, abs/1811.00982.

\bibitem[Li et~al., 2020]{li2020fast}
Li, Y., Luo, A., and Lyu, S. (2020).
\newblock Fast portrait segmentation with highly light-weight network.
\newblock In {\em 2020 IEEE International Conference on Image Processing
  (ICIP)}, pages 1511--1515. IEEE.

\bibitem[Liang et~al., 2022]{Liang2022}
Liang, Z., Guo, K., Li, X., Jin, X., and Shen, J. (2022).
\newblock Person foreground segmentation by learning multi-domain networks.
\newblock {\em IEEE Transactions on Image Processing}, 31:585--597.

\bibitem[Lin et~al., 2017a]{lin2017feature}
Lin, T.-Y., Doll{\'a}r, P., Girshick, R., He, K., Hariharan, B., and Belongie,
  S. (2017a).
\newblock Feature pyramid networks for object detection.
\newblock In {\em Proceedings of the IEEE conference on computer vision and
  pattern recognition}, pages 2117--2125.

\bibitem[Lin et~al., 2017b]{lin2017focal}
Lin, T.-Y., Goyal, P., Girshick, R., He, K., and Doll{\'a}r, P. (2017b).
\newblock Focal loss for dense object detection.
\newblock In {\em Proceedings of the IEEE international conference on computer
  vision}, pages 2980--2988.

\bibitem[Lin et~al., 2014]{coco}
Lin, T.-Y., Maire, M., Belongie, S., Bourdev, L., Girshick, R., Hays, J.,
  Perona, P., Ramanan, D., Zitnick, C.~L., and Dollár, P. (2014).
\newblock Microsoft coco: Common objects in context.
\newblock cite arxiv:1405.0312Comment: 1) updated annotation pipeline
  description and figures; 2) added new section describing datasets splits; 3)
  updated author list.

\bibitem[Orts-Escolano and Ehman, 2022]{google_alpha_matting}
Orts-Escolano, S. and Ehman, J. (2022).
\newblock {Accurate Alpha Matting for Portrait Mode Selfies on Pixel 6 }.
\newblock
  \url{https://ai.googleblog.com/2022/01/accurate-alpha-matting-for-portrait.html}.

\bibitem[Park et~al., 2019]{park2019extremec3net}
Park, H., Sj{\"o}sund, L.~L., Yoo, Y., and Kwak, N. (2019).
\newblock Extremec3net: Extreme lightweight portrait segmentation networks
  using advanced c3-modules.
\newblock {\em arXiv preprint arXiv:1908.03093}.

\bibitem[Smith et~al., 2019]{smith2019towards}
Smith, B.~M., Chari, V., Agrawal, A., Rehg, J.~M., and Sever, R. (2019).
\newblock Towards accurate 3d human body reconstruction from silhouettes.
\newblock In {\em 2019 International Conference on 3D Vision (3DV)}, pages
  279--288. IEEE.

\bibitem[Song et~al., 2016]{song20163d}
Song, D., Tong, R., Chang, J., Yang, X., Tang, M., and Zhang, J.~J. (2016).
\newblock 3d body shapes estimation from dressed-human silhouettes.
\newblock In {\em Computer Graphics Forum}, volume~35, pages 147--156. Wiley
  Online Library.

\bibitem[Song et~al., 2018]{8360117}
Song, D., Tong, R., Du, J., Zhang, Y., and Jin, Y. (2018).
\newblock Data-driven 3-d human body customization with a mobile device.
\newblock {\em IEEE Access}, 6:27939--27948.

\bibitem[Strohmayer et~al., 2021]{strohmayer2021efficient}
Strohmayer, J., Knapp, J., and Kampel, M. (2021).
\newblock Efficient models for real-time person segmentation on mobile phones.
\newblock In {\em 2021 29th European Signal Processing Conference (EUSIPCO)},
  pages 651--655. IEEE.

\bibitem[Tan et~al., 2019]{tan2019mnasnet}
Tan, M., Chen, B., Pang, R., Vasudevan, V., Sandler, M., Howard, A., and Le,
  Q.~V. (2019).
\newblock Mnasnet: Platform-aware neural architecture search for mobile.
\newblock In {\em Proceedings of the IEEE/CVF Conference on Computer Vision and
  Pattern Recognition}, pages 2820--2828.

\bibitem[Wu et~al., 2019a]{wu2019fbnet}
Wu, B., Dai, X., Zhang, P., Wang, Y., Sun, F., Wu, Y., Tian, Y., Vajda, P.,
  Jia, Y., and Keutzer, K. (2019a).
\newblock Fbnet: Hardware-aware efficient convnet design via differentiable
  neural architecture search.
\newblock In {\em Proceedings of the IEEE/CVF Conference on Computer Vision and
  Pattern Recognition}, pages 10734--10742.

\bibitem[Wu et~al., 2015]{wu2015quantized}
Wu, J., Leng, C., Wang, Y., Hu, Q., and Cheng, J. (2015).
\newblock Quantized convolutional neural networks for mobile devices. arxiv
  eprints, page.
\newblock {\em arXiv preprint arXiv:1512.06473}.

\bibitem[Wu et~al., 2019b]{wu2019detectron2}
Wu, Y., Kirillov, A., Massa, F., Lo, W.-Y., and Girshick, R. (2019b).
\newblock Detectron2.
\newblock \url{https://github.com/facebookresearch/detectron2}.

\bibitem[Xi et~al., 2019]{Xi2019}
Xi, W., Chen, J., Lin, Q., and Allebach, J.~P. (2019).
\newblock High-accuracy automatic person segmentation with novel spatial
  saliency map.
\newblock In {\em 2019 IEEE International Conference on Image Processing
  (ICIP)}, pages 1560--1564.

\bibitem[Xie et~al., 2017]{xie2017aggregated}
Xie, S., Girshick, R., Doll{\'a}r, P., Tu, Z., and He, K. (2017).
\newblock Aggregated residual transformations for deep neural networks.
\newblock In {\em Proceedings of the IEEE conference on computer vision and
  pattern recognition}, pages 1492--1500.

\bibitem[Yan et~al., 2021]{yan2021silhouette}
Yan, S., Wirta, J., and K{\"a}m{\"a}r{\"a}inen, J.-K. (2021).
\newblock Silhouette body measurement benchmarks.
\newblock In {\em 2020 25th International Conference on Pattern Recognition
  (ICPR)}, pages 7804--7809. IEEE.

\end{thebibliography}


\end{document}